\documentclass[10pt,twocolumn,letterpaper]{article}

\usepackage{cvpr}
\usepackage{times}
\usepackage{epsfig}
\usepackage{graphicx}
\usepackage{amsmath}
\usepackage{amssymb}

\DeclareMathOperator*{\argmin}{arg\,min}
\usepackage[ruled]{algorithm2e}
\usepackage{booktabs}
\usepackage{multirow}

\usepackage[pagebackref=true,breaklinks=true,letterpaper=true,colorlinks,bookmarks=false]{hyperref}

\cvprfinalcopy % *** Uncomment this line for the final submission

\def\cvprPaperID{1073} % *** Enter the CVPR Paper ID here

\ifcvprfinal\fi
\begin{document}
%%%%%%%%% TITLE
\title{Rethinking Performance Estimation in Neural Architecture Search}

\author{
    Xiawu Zheng $^{1,2,3}$\ ,
    Rongrong Ji$^{1,2,3}$\thanks{Corresponding Author.} ,
    Qiang Wang$^{1,3}$,
    Qixiang Ye$^{3,4}$,
    Zhenguo Li$^5$
    Yonghong Tian$^{3,6}$,
    Qi Tian$^5$ \\
    $^1$Media Analytics and Computing Lab, Department of Artificial Intelligence,\\
    School of Informatics, Xiamen University, 361005, China\\
    $^2$National Institute for Data Science in Health and Medicine, Xiamen University.\\
    $^3$Peng Cheng Laboratory, Shenzhen, China.
    $^4$University of Chinese Academy of Sciences, China.\\
    $^5$Noah's Ark Lab,  Huawei Technologies $^6$Peking University.\\
    {\tt\small \{zhengxiawu,Wangqiang\}@stu.xmu.edu.cn, rrji@xmu.edu.cn}\\
    {\tt\small qxye@ucas.ac.cn, li.zhenguo@huawei.com,yhtian@pku.edu.cn, tian.qi1@huawei.com}
}

\maketitle
\thispagestyle{empty}

%%%%%%%%% ABSTRACT
\begin{abstract}
%Neural architecture search (NAS) has been recently popular in building deep models for computer vision applications and beyond.
%
%Despite of gexciting progress, most NAS methods remain limited by practical resource constraints, mainly due to the computationally expensive step of performance estimation (PE), i.e., iteratively estimating the performance of searched network structure. 
Neural architecture search (NAS) remains a challenging problem, which is attributed to the indispensable and time-consuming component of performance estimation (PE).
%which iteratively estimate the performance of searched network structure.
%
%owever, in the NAS community, few works have been devoted to deeply investigating this issue.
%
In this paper, we provide a novel yet systematic rethinking of PE in a resource constrained regime, termed budgeted PE (BPE), which precisely and effectively estimates the performance of an architecture sampled from an architecture space.
%BPE is essentially a group of optimal hyper-parameters (channel, layers, learning rate and image size etc).}
%
Since searching an optimal BPE is extremely time-consuming as it requires to train a large number of networks for evaluation, we propose a Minimum Importance Pruning (MIP) approach.
Given a dataset and a BPE search space, MIP estimates the importance of hyper-parameters using random forest and subsequently prunes the minimum one from the next iteration. 
%aims to find an optimal BPE, which solves {\color{blue}above problems} by 
%这带来的优点
In this way, MIP effectively prunes less important hyper-parameters to allocate more computational resource on more important ones, thus achieving an effective exploration.
By combining BPE with various search algorithms including reinforcement learning, evolution algorithm, random search, and differentiable architecture search, we achieve $1,000 \times$ of NAS speed up with a negligible performance drop comparing to the SOTA.\footnote{All the NAS search codes are available at: \url{https://github.com/zhengxiawu/rethinking_performance_estimation_in_NAS}}

%obtain one performance rank for validation, which is fully unexploited in the NAS community.
%
%the best achievable
%
%We validate our method through extensive experiments 
\end{abstract}

%%%%%%%%% BODY TEXT
\section{Introduction}\label{sec:intro}

\begin{figure}
\includegraphics[width=1.0\linewidth]{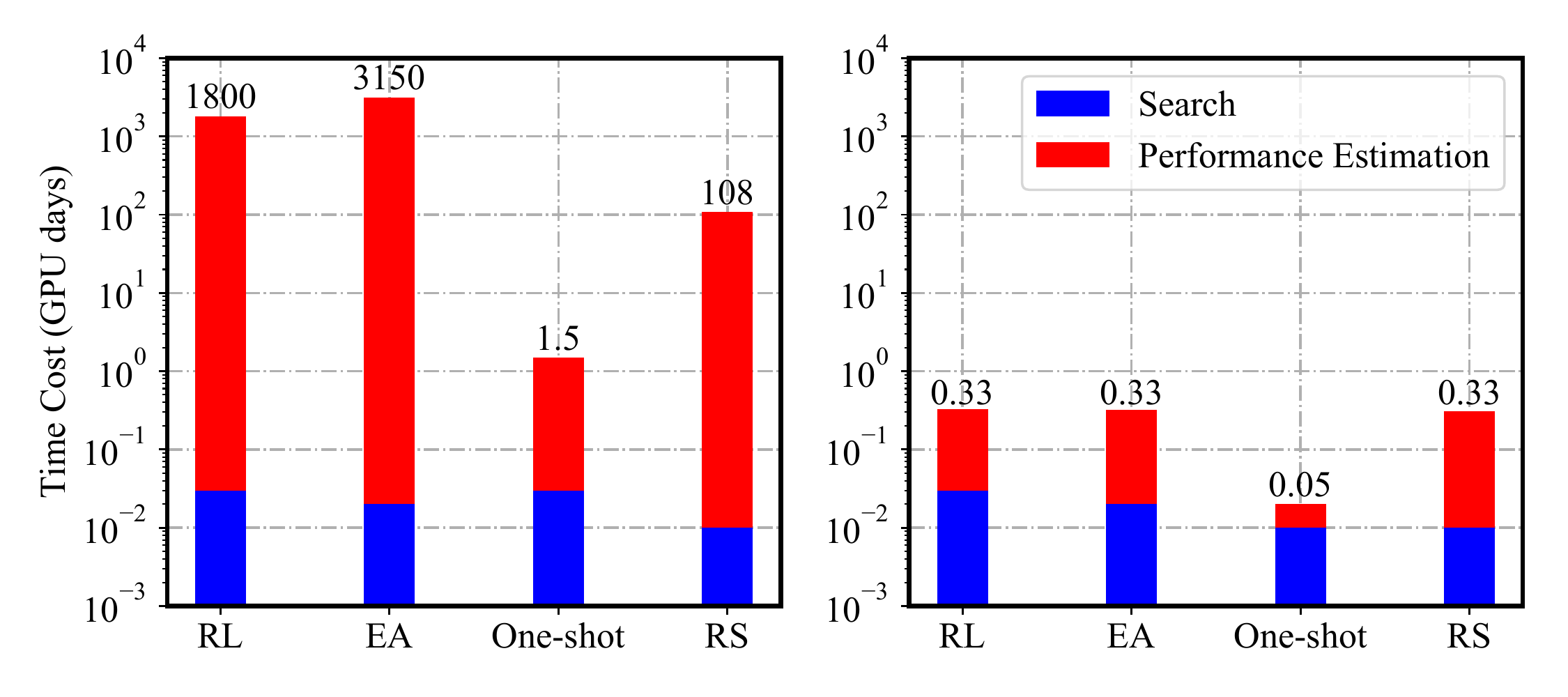}
\caption{\label{fig:time_cost} Time cost with/without budgeted performance estimation (BPE-1). 
(a) Previous methods did not optimize the huge computation cost in PE. 
(b) By incorporating the BPE, we can largely accelerate NAS methods including reinforcement learning (RL) \cite{zoph2018learning}, evolution algorithm (EA) \cite{real2018regularized}, random search (RS) and DARTS \cite{liu2018darts} (One-shot) with a negligible performance drop. % Detailed evaluations can be found Sec.~\ref{sec:5}.
}
\end{figure}

\begin{figure*}
\centering
\includegraphics[width=0.9\linewidth]{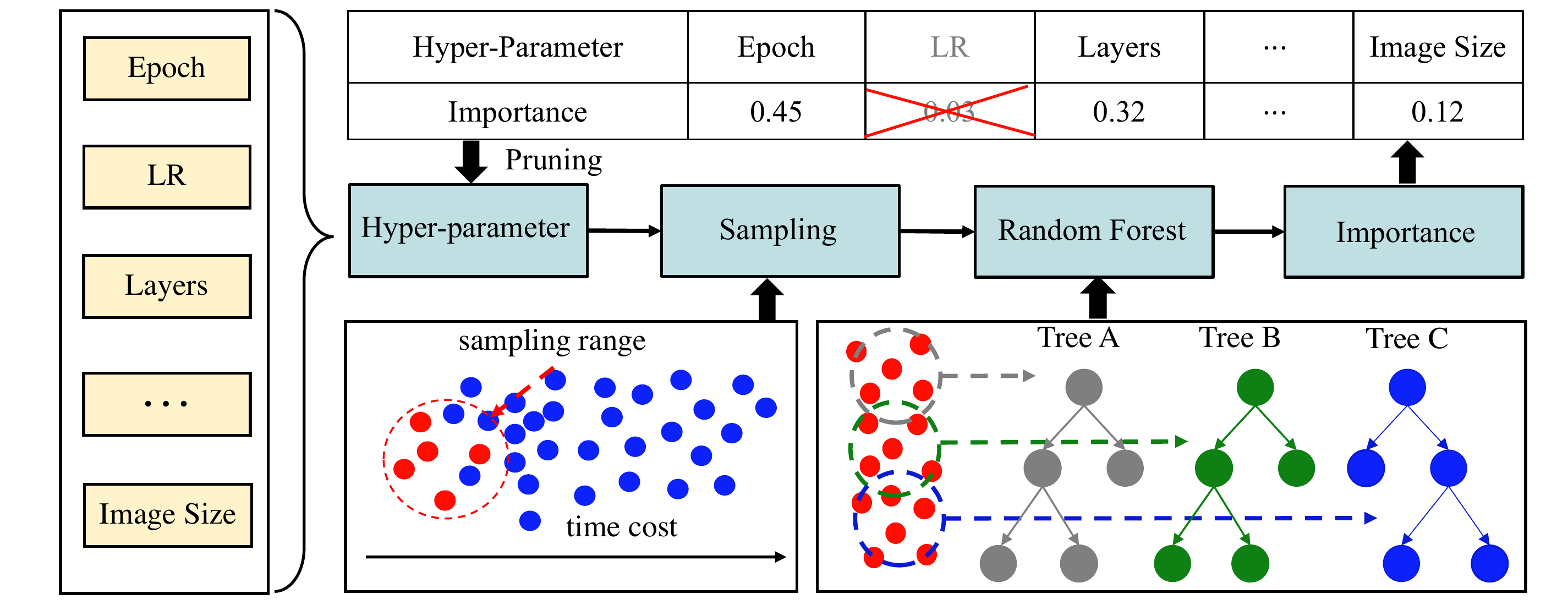}
\caption{\label{fig:framework} 
The overall framework of the proposed Minimum Importance Pruning for finding an optimal Budgeted Performance Estimation. 
The search space of BPE is built from the training hyper-parameters including training epoch, batch size, learning rate, layer number, float point, channels, cutout and image size. 
We first sample the example with the lowest time cost. 
Then the sampled example is used to train the random forest, which is used to evaluate the importance of the corresponding hyper-parameters. 
%
%write
The hyper-parameter with the lowest importance is pruned by assigning the value with a minimum time cost.}
\end{figure*}

Deep learning have made significant sucess in classification \cite{he2016deep,Cheng_2019_CVPR}, retrieval \cite{zheng2019towards, zheng2018centralized} and detection \cite{CMIL2019,FreeAnchor2019,ji2019semi}. To this end, neural architecture search (NAS) aims to automatically discover a suitable neural network architecture by exploring over a tremendous architecture search space, which has shown remarkable performance over manual designs in various computer vision tasks \cite{chen2019binarized, zoph2016neural,liu2018darts,zoph2018learning,chen2018searching,liu2019auto}. 

Despite the extensive success, previous methods are still defective in intensive computation resources, which severely restricts its application prospect and flexibility. 
For instance, reinforcement learning (RL) based methods \cite{zoph2018learning, zoph2016neural} search a suitable architecture on CIFAR10 by training and evaluating more than $20,000$ architectures by using $500$ GPUs over $4$ days. For another instance, the evolutionary algorithm (EA) based method in  \cite{real2018regularized} needs $3,150$ GPU days to find an optimal architecture on CIFAR10.

A NAS method generally consists of three components, \emph{i.e.}, search space, search strategy and performance estimation. 
As established by \cite{zoph2016neural}, cell based search space is now well adopted \cite{zheng2019dynamic, xu2019pc, chen2019progressive, pham2018efficient,real2018regularized,real2019aging,zoph2016neural,zoph2018learning}, which is pre-defined and fixed during the architecture search to ensure a fair comparison among different NAS methods. 
On the other hand, as illustrated in Fig.~\ref{fig:time_cost}, different search strategies (RL or EA) have similar run-time (after subtracting the performance estimation cost), which can also be well accelerated with GPU packages. 
%
% Besides, as validated in \cite{li2019random,sciuto2019evaluating}, random search even provides very comparable performance to the state-of-the-arts like DARTS \cite{liu2018darts}.
%
Therefore the  major computational consumption of NAS lies in the performance estimation (PE) step, as validated in Fig.~\ref{fig:time_cost}. 
However, few works have been devoted to the efficiency issue of PE, which is crucial to cope with the explosive growth of dataset size and model complexity.
Moreover, it is highly desirable to conduct fast architecture search under different datasets for deployment in emerging applications like self-driving cars \cite{bojarski2016end}.

In this paper, we propose a novel and efficient performance estimation under the resource-constrained regime, termed \emph{budgeted performance estimation} (BPE), which is the first of its kind in the NAS community. 
%
%write
The BPE essentially controls the hyper-parameters of training, network designing and dataset processing, such as number of channels, number of layers, learning rate and image size.
Rather than pursuing model precision for a specific dataset, BPE aims to learn the most achievable \emph{relative precision order} of different neural architectures in a specific architecture space.
%这里的relative precision order加一个说明，值得是假如说一个网络空间里面有三个网络，在训练到收敛的情况下，排序关系是A>B>C,我们希望的是在找到某一组超参数下，只用一点点时间，我们就可以得出A>B>C的关系。
%
In other words, a good network structure still has a relatively high ranking on an accurate BPE.
We argue that the missing of accurate and efficient BPE remains as the main barrier for the wide usage of NAS research.
However, finding an accurate and effective BPE is extremely challenging compared to other black-box optimization problems. 
First, BPE needs to carefully deal with the discrete (like layers or channels) and continuous (like learning rate) hyper-parameters. 
Second, evaluating a specific BPE needs to train a large number of neural networks \emph{e.g.}, $2\times8^{14}$ networks in the cell-based archietcture search space \cite{liu2018darts}.

As implicitly employed in previous NAS \cite{zoph2018learning,klein2016fast,real2019aging, liu2018darts, cai2018efficient,pham2018efficient} methods, most BPE methods only leverage intuitive tricks including early stopping \cite{zoph2018learning}, dataset sampling \cite{klein2016fast} and lower resolution dataset, or using a proxy search network with fewer filters per layer and fewer cells \cite{zoph2018learning, liu2018darts}. 
While such methods can reduce the computational cost to a certain extent (which is still time consuming \cite{zoph2018learning, real2018regularized}), noise is also introduced into PE to underestimate its corresponding performance. 
Little work investigates the relative performance rank between approximated evaluations and full evaluations, which is traditionally considered as a merited trick \cite{liu2018darts, zoph2018learning, real2018regularized}.
However, as subsequently validated in Sec.~\ref{sec:5}, such a relative rank can change dramatically under a tiny difference in the training condition.

In this paper, we present a unified, fast and effective framework, termed
Minimum Importance Pruning (MIP), to find an optimal BPE on a specific architecture search space such as cell-based search space \cite{zheng2019dynamic, xu2019pc, chen2019progressive, pham2018efficient,real2018regularized,real2019aging,zoph2016neural,zoph2018learning}, as illustrated in Fig.~\ref{fig:framework}. 
In particular, for a given large-scale hyper-parameter search space, we first sample examples with the lowest time consumption. 
 The sampled examples are then used to estimate the hyper-parameter importance using random forest \cite{breiman2001random, hoos2014efficient}. 
The hyper-parameter of the lowest importance is set to the value with the minimum time cost. 
The algorithm stops when every hyper-parameter is set. 
%We validate the search efficiency and network performance by combining the found BPE with various search algorithms, including Reinforcement Learning (RL), Evolutionary Algorithms (EA), Random Search (RS) and One-shot based method. 
%
%Our method reaches the state-of-the-art test error on CIFAR-10 \cite{krizhevsky2010convolutional}  (\emph{i.e.}, $2.44$ \%), while the search process is more than a magnitude times faster. 
%
%On ImageNet \cite{russakovsky2015imagenet}, our model achieves $74.9$\% top-1 accuracy under the MobileNet settings (MobileNet V1/V2 \cite{howard2017mobilenets, sandler2018mobilenetv2}). 
%
%
The contributions of this paper include:
\begin{itemize}
\item It is the first work to systematically investigate the performance estimation in NAS under the resource-constrained regime. 
We seek an optimal budgeted PE (BPE) by designing a spearman correlation loss function on a group of key hyper-parameters. 

%write
\item A novel hyper-parameter optimization method, termed Minimum Importance Pruning (MIP), is proposed, which is effective for black-box optimization with extremely time consuming on the evaluation step. 
%The proposed MIP can automatically and effectively find the optimal BPE in the widely-used cell based NAS search spaces.
%\item We make the first effort to understand the one-shot based methods such as DARTS \cite{liu2018darts} by using the found BPE. In Sec .\ref{sec:5}, we found that the effectiveness and randomness are primarily because corresponding PE is accurate in a small range.
%Differing from the existing works like \cite{sciuto2019evaluating} that attribute such methods to negatively impact the rank order through the whole search space, 

\item The proposed MIP-BPE generalizes well to various architecture search methods, including Reinforcement Learning (RL), Evolutionary Algorithms (EA), Random Search (RS) and DARTS. 
MIP-BPE  achieves remarkable performance on both CIFAR-10 and ImageNet, while accelerating the search process by $1,000 \times$.
\end{itemize}

\section{Related Work}
\subsection{Performance Estimation in NAS}
Performance estimation refers to estimating the performance of a specific architecture in the architecture search space. 
A conventional option is to perform a standard training and validation process of this architecture on the dataset, which is computationally expensive and limits the number of architectures that can be explored. 
To accelerate performance estimation, most NAS methods only provide simple intuitive cues such as early stopping \cite{zoph2018learning}, dataset sampling \cite{klein2016fast} and lower resolution dataset, or using a proxy search network with fewer filters and fewer cells \cite{zoph2018learning, liu2018darts}. 

Another possibility of estimating the architecture performance is one-shot based methods  \cite{Zheng_2019_ICCV, liu2018darts, AkimotoICML2019}, which consider each individual in the search space as a sub-graph sampled from a super-graph.
In this way, they accelerate the search process by parameter sharing \cite{pham2018efficient}.
Chen \emph{et al.} \cite{chen2019progressive} proposed to progressively grow the depth of searched architectures during the training procedure. 
Xu \emph{et al.} \cite{xu2019pc} presented a partially connected method by sampling a small part of the super-net to reduce the redundancy in the network space, which thereby performs a more efficient search without comprising the performance. 
However, these methods do not deeply investigate the influence of different hyper-parameters, which has introduced large noise as validated in Sec.~\ref{sec:5}.

\subsection{Hyper-parameter Optimization}
Hyper-parameter optimization \cite{thornton2013auto} aims to automatically optimize the hyper-parameters during the learning process \cite{bergstra2012random, hutter2011sequential, snoek2012practical, zela2018towards}.
To this end, gird search and random search \cite{bergstra2012random} are the two simplest and most straightforward approaches. 
 Note that these methods do not consider to use the experience (sampled examples in the search process).
Subsequently, sequential model-based optimization (SMBO) \cite{hutter2011sequential} is proposed to learn a proxy function from the experience and estimate the performance for unknown hyper-parameters.
As one of the most popular methods, Bayesian optimization \cite{snoek2012practical} learns a Gaussian process with the sampled examples, and then decides the best hyper-parameter for the next trial by maximizing the corresponding improvement function.

 However, all these methods mostly deal with hyper-parameters for particular machine learning models, which cannot handle the optimization of BPE with such an expensive evaluation step.
 Different from the previous methods, we evaluate and estimate the importance of the hyper-parameters by sampling examples with the minimum time consumption, where hyper-parameters of minimum importances are then pruned in the next iteration, which is extremely effective and efficient to find the optimal BPE.
\section{Preliminaries}
\subsection{NAS Pipeline} \label{sec: 3.1}
Given a training set, conventional NAS algorithms \cite{zoph2018learning, zheng2019dynamic, li2019random} first sample an architecture in the pre-defined search space by a certain search strategy like Reinforcement Learning (RL) or Evolution Algorithm (EA). 
Then the sampled neural architecture is passed to the performance estimation (PE), which returns the performance of the architecture to the search algorithm.

In most NAS methods \cite{zheng2019dynamic, liu2018darts,xu2019pc}, PE is accelerated by using a group of lower-cost hyper-parameters (like smaller image size, less channel and shallower network) in the search space $\Omega = \Theta_1 \times \Theta_2 \times ...\times \Theta_n$, termed budgeted PE (BPE), which contains $n$ sorts of training hyper-parameters including the number of training epochs, batch size, learning rate, the number of layers, float point precision, channels, cutout \cite{devries2017improved} and image size.
For instance, Liu \emph{et al.} \cite{liu2018darts} proposed to estimate the performance of an architecture on a small network of $8$ layers trained for $50$ epochs, with batch size $64$ and initial number of channels $16$.
After the search process, the optimal neural architecture is then evaluated by a \emph{fully and time-consuming training hyper-parameter} set $f$. 
In the existing works \cite{zheng2019dynamic, liu2018darts,xu2019pc}, $f$ controls the final evaluation hyper-parameters of the optimal architecture, \emph{i.e.,} a large network of $20$ layers is trained for $600$ epochs with a batch size of $96$ and an additional regularization such as cutout \cite{devries2017improved}. 

However, in this pipeline, the BPE and the final evaluation phase are decoupled. 
There is no guarantee that the BPE is correlated to the final evaluation step, \emph{i.e.,} the same architectures may have large ranking distances under different training conditions. 
Most NAS methods \cite{liu2018darts, zoph2018learning} intuitively change BPE with fewer channels or layers. 
Nevertheless, extensive experiments in Sec.~\ref{sec:5} show that the effectiveness of BPE is very sensitive, which means that it needs to carefully select and analyze the corresponding hyper-parameters in NAS. 
Indeed, we believe, and validated in Sec. \ref{sec:5}, that BPE is a crucial component, while unfortunately there are no corresponding works devoted to this area. 
\subsection{Cell based Architecture Search Space} \label{subsec: 3.2}
As mentioned in Sec.~\ref{sec:intro}, BPE aims to find optimal training hyper-parameters on a specific architecture search space. 
In this paper, we follow the widely-used cell-based architecture search space in \cite{zheng2019dynamic, xu2019pc, chen2019progressive, pham2018efficient,real2018regularized,real2019aging,zoph2016neural,zoph2018learning, Zheng_2019_ICCV}:
A network consists of a pre-defined number of cells \cite{zoph2016neural}, which can be either norm cells or reduction cells. Each cell takes the outputs of the two previous cells as input. 
A cell is a fully-connected directed acyclic graph (DAG) of $M$ nodes, \emph{i.e.}, $\{B_1, B_2, ..., B_M\}$.
Each node $B_i$ takes the dependent nodes as input, and generates an output through a sum operation $B_j = \sum_{i<j} o^{(i,j)}(B_i).$ 
Here each node is a specific tensor (\emph{e.g.,} a feature map in convolutional neural networks) and each directed edge $(i,j)$ between $B_i$ and $B_j$ denotes an operation $o^{(i,j)} (.)$, which is sampled from the corresponding operation search space $\mathcal{O}^{(i,j)}$. 
Note that the constraint $i<j$ ensures no cycles in a cell. 
Each cell takes the outputs of two dependent cells as input, and the two input nodes are set as $B_{-1}$ and $B_{0}$ for simplicity. 
Following \cite{liu2018darts}, the operation search space $\mathcal{O}^{(i,j)}$ consists of $K = 8$ operations: $3\times3$ dilated convolution with rate $2$, $5\times5$ dilated convolution with rate $2$, $3\times3$ depth-wise separable convolution,  $5\times5$ depth-wise separable convolution, $3\times3$ max pooling, $3\times3$ average pooling, no connection (zero), and a skip connection (identity). 
Therefore, the size of the whole search space $\mathcal{O}$ is $2 \times K^{|\mathcal{E_M}|}$, where $\mathcal{E_M}$ is the set of possible edges with $M$ intermediate nodes in the fully-connected DAG. 
In our case with $M=4$ the total number of cell structures in the search space is $2 \times 8^{2+3+4+5} = 2 \times 8^{14}$, which is an extremely large space to search.

\section{The Proposed Method}
In this section, we first describe the formal setting of BPE in Sec.~\ref{sec:4.1}.  We then present the proposed minimum importance pruning (MIP) to find the optimal BPE. 
\subsection{Budgeted Performance Estimation}\label{sec:4.1}
The performance estimation is a training algorithm $A$ with $n$ hyper-parameters in a domain $\Omega$.
%
%write
Given an architecture set $\mathcal{G}$ sampled from $\mathcal{O}$, we address the following optimization problem:
\begin{equation}\label{eq:objective}
\max_{b\in \Omega}\, r_s(R_f, R_b) + \lambda \overline{T(\mathcal{G}_b)},
\end{equation}
where $0< \lambda <1$, $r_s$ calculates the Spearman Rank Correlation between $R_f$ and $R_b$. 
$R_f$ and $R_b$ are the performance on validation set of every architecture in $\mathcal{G}$ with full training hyper-parameter $f$ and BPE parameter $b$, respectively. 
We aim to find the optimal $b$ with less average training consumption $\overline{T(\mathcal{G}_b)}$ on $\mathcal{G}$. 

Optimizing Eq.~\ref{eq:objective} is extremely challenging, as we need to train over $|\mathcal{G}|$ architectures to validate one example in $\Omega$. 
This large set of models to be trained and evaluated prevent most NAS methods to be widely deployed. 
Fortunately, Radosavovic \emph{et al.} \cite{radosavovic2019network} observed that sampling about $100$ models form a given architecture search space is sufficient to perform robust estimation, which is also validated in our work. 
Specifically, we randomly sample $100$ neural architectures in the cell-based search architecture space to construct the architecture set $\mathcal{G}$. 
Then $R_f$ and $R_b$ are obtained by training and validate every architecture in $\mathcal{G}$ with the hyper-parameters $f$ and $b$, respectively.

\subsection{Minimum Importance Pruning}\label{sec:4.2}
Although the time consumption of the validation step has been drastically reduced, it is still very difficult to optimize Eq.~\ref{eq:objective}, \emph{i.e.,} in our evaluation, the average training time of an architecture from $\mathcal{G}$ for different hyper-parameters is $10$ hours on CIFAR10 benchmark. 
In this case, each example needs to train $100$ networks, and the time consumption for one BPE example $b$ is $\sim 10^3$ hours. 
Such a time consumption is still difficult for finding an optimal BPE efficiently.  

To handle this issue we propose a minimum importance pruning (MIP) as illustrated in Fig.~\ref{fig:framework}. 
We first sample the hyper-parameter examples around the lowest time cost. 
Then the sampled examples are trained to estimate the hyper-parameter importance by using random forest \cite{hoos2014efficient, breiman2001random}.  
After that, the hyper-parameter with the lowest importance is pruned by setting the value with the minimum time cost. 
The pruning step is ceased when there is only one hyper-parameter in the search space, and the optimal BPE is the example with the maximum $r_s$. 

\textbf{Lowest time cost sampling.} 
For each element in $\Theta_i$, we introduce a category distribution related to the computational cost: 
\begin{equation}\label{eq:sapling}
p(\eta_{i,j}) = \frac{\exp\{-FLOPs(\Theta_{i,j})\}}{\sum_{j}{\exp\{-FLOPs(\Theta_{i,j})\}}},
\end{equation}
where $\Theta_{i,j}$ denotes the $j$th element of the $i$th hyper-parameter $\Theta_i$. 
The function $FLOPs(\Theta_{i,j})$ is the number of floating point operations. We set $\Theta_i$ with the $j$th element and fix other hyper-parameters in $\Omega$ by the value with the minimum time cost. 
An example $b$ is generated by sampling the joint probability in Eq.~\ref{eq:sapling}, \emph{e.g.,} $p(\eta) = \prod_{i=1}^n p(\eta_i)$. 
Then, we obtain $R_b$ by training every architecture in $\mathcal{G}$ using the sampled $b$, and the objective $r_s$ is calculated with $R_f$ by using Eq.~\ref{eq:objective}.
\begin{algorithm}[t]
\small
\caption{Minimum Importance Pruning \label{alg:MIP}}
\LinesNumbered
\KwIn{Architecture search space $\mathcal{O}$; hyper-parameter space $\Omega$; sampling time $K=10$}
\KwOut{Optimal BPE hyper-parameter $b$.}

$n \leftarrow$ The number of parameter in $\Omega$;\\
$R \leftarrow$ \{Sample $100$ network architectures in $\mathcal{O}$\};\\
Train $R$ with fully training condition $f$;\\
$R_f \leftarrow$ \{$R$'s performance on validation set\};\\
$\mathcal{D} \leftarrow \emptyset$;\\
\While{($n>0$) }{
$B = $\{Randomly sample $T$ examples in $\Omega$ by distribution in Eq.~\ref{eq:sapling} \};\\

\For{$b$ in $B$}{
Train $R$ with BPE $b$;\\
$R_b \leftarrow$ \{$R$'s performance on validation set\};\\
$r_s \leftarrow$ Spearman Rank Correlation between $R_b$ and $R_f$;\\
$\mathcal{D} \leftarrow \mathcal{D} \cup \{b, r_s\}$
}
 Train random forest $\mathcal{T}$ by using Eq.~\ref{eq:gini} and Eq.~\ref{eq:partition} on $\mathcal{D}$;\\
 Calculate the importance by Eq.~\ref{eq:parameter_importance};\\
 Pruning space $\Omega$ by Eq.~\ref{eq:pruning};\\
$n = n -1$;
}
\end{algorithm}

\textbf{Random forest training.} 
After repeating previous steps over $K=10$ times, we get a set $\mathcal{D} = \{(b_1, r_{s,1}), (b_2, r_{s,2}), ...,(b_K, r_{s,K})\}$ with different BPEs and corresponding objective values, which is used as a training set for the random forest.  
In random forest, each tree is built from a set $\mathcal{D}_s$ drawn with a replacement sampling from $\mathcal{D}$. 
Training random forest is to train multiple regression trees. 
Given a training set $\mathcal{D}_s$ with $b_i \in \mathbb{R}^n, i = 1,...,l$ and the corresponding spearman rank correlation vector $[r_{s,1}, r_{s,2},..., r_{s,l}]$ sampled from $\mathcal{D}$, a regression tree in the random forest recursively partitions the space such that the examples in $\mathcal{D}_s$ with similar values are grouped together. 
When training the regression tree, we need to consider how to measure and choose the partition feature (hyper-parameter in our case).  
Specifically, let the data at node $m$ be represented by $Q$. For each candidate partition $\xi = (i,t_m)$ consisting of hyper-parameter $i$ and threshold $t_m$, we partition the data into $Q_{left}(\xi)$ and $Q_{right}(\xi)$ subsets as follows:
\begin{equation}
\small
\begin{split}
&Q_{left}(\xi) = (x, r_s)\,|x_i \leq t_m\\
&Q_{right}(\xi) = (x, r_s)\,|x_i > t_m.
\end{split}
\end{equation}
We further define the impurity function $H(\cdot)$ for a given split set $Q$ as
\begin{equation}
\small
H\left( Q\right) = \frac{1}{|Q|} \sum_{r_{s,i}\in Q}\left(r_{s,i}-\overline{r_{s, Q}}\right)^2,
\end{equation}
where $\overline{r_{s, Q}} = \frac{1}{|Q|} \sum_{r_{s,i}\in Q} r_{s,i}$, $|Q|$ denotes the number of examples in set $Q$. 
And the impurity for a specific partition is the weighted sum of the impurity function:
\begin{equation}\label{eq:gini}
\small
G(Q, \xi) = \frac{|Q_{left}|H(Q_{left})+ |Q_{right}|H(Q_{right})}{|Q_{left}|+|Q_{right}|}.
\end{equation}
We adopt the exhaustion method to find the optimal partition, that is, iterate through all possible partitions and select the partition with the minimum impurity:
\begin{equation}\label{eq:partition}
\xi^* = \argmin_{\xi} G(Q, \xi).
\end{equation} 

\textbf{Hyper-parameter importance.} 
For every node $m$ in the regression tree, we calculate the parameter importance as the decrease in node impurity, which is weighted by the number of samples that reach the node. 
The parameter importance for node $m$ is defined as:
\begin{equation}
\small
\begin{split}
I_m = & |Q_m|H(Q_m) - |Q_{\{left,m\}}|H\left( Q_{\{left,m\}}\right)\\
&  - |Q_{\{right,m\}}|H\left( Q_{\{right,m\}}\right).
\end{split}
\end{equation}
The importance for each $\Theta_i$ is the summation of the importance through the node in the random forest, which uses $\Theta_i$ as the partition parameter:
\begin{equation}\label{eq:parameter_importance}
I_{\Theta_i} = \frac{\sum_{\xi_m(0) = \Theta_i}I_m}{\sum I_m}.
\end{equation}

\textbf{Parameter pruning.} 
After the importance estimation process in Eq.~\ref{eq:parameter_importance}, the hyper-parameter with the lowest probability is pruned by setting
\begin{equation}\label{eq:pruning}
\Theta_i = \beta_i, \, \Theta_i  = \argmin I_{\Theta}.
\end{equation}
$\beta_i$ is the value of the lowest FLOPs in hyper-parameter $\Theta_i$ when $I_{\Theta_i} < 0.1$. Otherwise, $\beta_i$ is the corresponding parameter value with the maximum $r_s$ in $\mathcal{D}$.  
The pruning step significantly improves the search efficiency. By setting the less important hyper-parameter to a value with less resource consumption, we can allocate more computational resource on important parameters. 
Our minimum importance pruning algorithm is presented in Alg.~\ref{alg:MIP}.

\begin{table}[t]
\begin{center}
\small
\begin{tabular}{lccc}
\hline
\textbf{Hyper-parameter}  & \textbf{BPE-1} & \textbf{BPE-2} & \textbf{DARTS}\cite{liu2018darts}\\
\hline
Epoch & 10 & \textbf{30}  & 50\\
Batch size & 128 & \textbf{128} & 64\\
Learning rate & 0.03 & \textbf{0.03} & 0.025\\
N\_Layers & 6 & \textbf{16} & 8\\
Channels & 8 & \textbf{16} & 16\\
Image Size & 16 & \textbf{16} & 32\\
Correlation $r_s$ & 0.50 & \textbf{0.63} & 0.57\\
Training Time & 0.08 & \textbf{0.55} & 1.38\\
\hline
\end{tabular}
\end{center}
\caption{Detailed hyper-parameters of the best settings discovered On CIFAR10 by using MIP. The found BPE-1 and BPE-2 show a better correlation $r_s$ with less average training time (GPU Hours).}
\label{tab:found_BPE}
\end{table}

\begin{table*}[t]
\small
\begin{center}
\setlength{\tabcolsep}{3.5mm}{
\begin{tabular}{lcccc}
\toprule[1pt]
\multirow{2}{*}{\textbf{Architecture}} & \textbf{Test Error} & \textbf{Params} & \textbf{Search Cost} & \textbf{Search} \\
& \textbf{(\%)} & \textbf{(M)} & \textbf{(GPU days)} & \textbf{Method} \\ \hline
ResNet-18 \cite{he2016deep} & 3.53 & 11.1 & - & Manual \\
DenseNet \cite{huang2017densely} & 4.77 & 1.0 & - & Manual \\
SENet \cite{hu2018squeeze} & 4.05 & 11.2 & - & Manual \\ \hline
NASNet-A \cite{zoph2018learning} & 2.65 & 3.3 & 1800 & RL \\
ENAS \cite{pham2018efficient} & 2.89 & 4.6 & 0.5 & RL \\
Path-level NAS \cite{cai2018path} & 3.64 & 3.2 & 8.3 & RL \\
\textbf{RL+BPE-1 (Ours)} & \textbf{2.66 $\pm$ 0.05} & \textbf{2.7} & \textbf{0.33} & \textbf{RL} \\
\textbf{RL+BPE-2 (Ours)} & \textbf{2.65 $\pm$ 0.12} & \textbf{2.9} & \textbf{2} & \textbf{RL} \\ \hline
AmoebaNet-B \cite{real2018regularized} & 2.55 & 2.8 & 3150 & Evolution \\
\textbf{EA+BPE-1 (Ours)} & \textbf{2.68 $\pm$ 0.09} & \textbf{2.46} & \textbf{0.33 } & \textbf{Evolution} \\ 
\textbf{EA+BPE-2 (Ours)} & \textbf{2.66 $\pm$ 0.07} & \textbf{2.87} & \textbf{2} & \textbf{Evolution} \\ \hline
DARTS \cite{liu2018darts} & 2.7 $\pm$ 0.01 & 3.1 & 1.5 & Gradient-based \\
GDAS \cite{Dong_2019_CVPR} & 2.93 $\pm$ 0.07 & 3.4 & 0.8 & Gradient-based \\
P-DARTS \cite{chen2019progressive} & 2.75 $\pm$ 0.06 & 3.4 & 0.3 & Gradient-based \\
SNAS \cite{xie2018snas} & 2.85 $\pm$ 0.02 & 2.8 & 1.5 & Gradient-based \\
\textbf{DARTS + BPE-1 (Ours)} & \textbf{2.89 $\pm$ 0.0} & \textbf{3.9} & \textbf{0.05} &  \textbf{Gradient-based} \\ 
\textbf{DARTS + BPE-2 (Ours)} & \textbf{2.72 $\pm$ 0.0} & \textbf{4.04} & \textbf{0.33} &  \textbf{Gradient-based} \\
\hline
Random Sample 100 & 2.55 & 2.9 & 108 & Random Search \\
\textbf{Random Sample 100 + BPE-1 (Ours)} & \textbf{2.68 $\pm$ 0.09} & \textbf{2.7} & \textbf{0.33(337$\times$)} & \textbf{Random Search}  \\
\textbf{Random Sample 100 + BPE-2 (Ours)} & \textbf{2.68 $\pm$ 0.05} & \textbf{1.9} & \textbf{2 (54$\times$)} & \textbf{Random Search}  

\\ \bottomrule[1pt]
\end{tabular}}
\end{center}
\caption{Comparing of test error rates for our discovered architecture, human-designed networks and other NAS architectures on CIFAR-10. For a fair comparison, we select the architectures and results with similar parameters ($<$ $5$M) and the same training condition (all the networks are trained with Cutout \cite{devries2017improved} ). Values are $\mu \pm \sigma$ across $4$ runs.}
\label{tab:cifar_results}
\end{table*}

\section{Experiment} \label{sec:5}

As we mentioned before, the average time consumption for evaluating BPE examples is $\sim 10^3$ GPU hours, which means that with similar sample magnitude ($\sim 70$), methods such as bayesian optimization or random search need about $7 \times 10^4$ GPU hours (almost infeasible). In contrast, our method needs only $5.2 \times 10^3$ GPU hours. Therefore, we do not compare these methods in our paper.

We first combine BPE with different search strategies including Reinforcement Learning (RL) \cite{kaelbling1996reinforcement},  Evolutionary Algorithm (EA) \cite{back1996evolutionary}, Random Search (RS) \cite{bergstra2012random} and Different Architecture Search (DARTS) \cite{liu2018darts}. 
As shown in Sec.~\ref{sec:compare_sota}, we compare with state-of-the-art methods in terms of both effectiveness and efficiency using CIFAR10 \cite{krizhevsky2009learning} and ImageNet \cite{russakovsky2015imagenet}. 
In Sec.~\ref{sec:analysis}, we investigate the effect of each hyper-parameter in BPE, as well as the efficiency of using Spearman Rank Correlation as the objective function. 
Although many works \cite{sciuto2019evaluating,li2019random} pointed out that the one-shot based method \cite{saxena2016convolutional, pham2018efficient, liu2018darts, bender2018understanding, Zheng_2019_ICCV} could not effectively estimate the performance throughout the entire search space,
in Sec.~\ref{sec:unstanding_one_shot} we have found that these methods are indeed effective in the \emph{local search space}, which reasonably explains the reproducibility and effectiveness, \emph{i.e.,} the corresponding algorithms are actually able to find good architectures, and the optimal architectures are quite different in different runs due to the local information.

\subsection{Comparing with State-of-the-arts} \label{sec:compare_sota}
We first search neural architectures by using the found BPE-1 and BPE-2 in Tab.~\ref{tab:found_BPE}, and then evaluate the best architecture with a stacked deeper network. 
To ensure the stability of the proposed method, we run each experiment $4$ times and find that the resulting architectures only show a slight variance in performance.

\subsubsection{Experimental Settings}
We use the same datasets and evaluation metrics for existing NAS methods \cite{liu2018darts,cai2018path,zoph2018learning,liu2018progressive}. 
First, most experiments are conducted on CIFAR-10 \cite{krizhevsky2010convolutional}, which has $50$K training images and $10$K testing images from 10 classes with a resolution $32 \times 32$. 
During the architecture search, we randomly select $5$K images from the training set as a validation set. 
To further evaluate the generalization capability, we stack the optimal cell discovered on CIFAR-10 into a deeper network, and then evaluate the classification accuracy on ILSVRC 2012 \cite{russakovsky2015imagenet}, which consists of $1,000$ classes with $1.28$M training images and $50$K validation images. 
Here, we consider the \emph{mobile} setting, where the input image size is $224 \times 224$ and the FLOPs is less than 600M.

In the search process, we directly use the found BPE-1 and BPE-2 in Tab.~\ref{tab:found_BPE} as the performance estimation with other search algorithms.
After finding the optimal architecture in the search space, we validate the final accuracy on a large network of $20$ cells is trained for $600$ epochs with a batch size of $96$ and additional regularization such as cutout \cite{devries2017improved}, which are similar to \cite{liu2018darts,zoph2018learning,pham2018efficient}.  
When stacking cells to evaluate on ImageNet, we use two initial convolutional layers of stride $2$ before stacking $14$ cells with the scale reduction at the $1$st, $2$nd, $6$th and $10$th cells. 
The total number of FLOPs is determined by the initial number of channels. The network is trained for 250 epochs with a batch size of 512, a weight decay of $3 \times 10 ^{-5}$, and an initial SGD learning rate of 0.1. 
All the experiments and models are implemented in PyTorch \cite{paszke2017automatic}.

\begin{table}[]
\small
\begin{center}
\setlength{\tabcolsep}{0.4mm}{
\begin{tabular}{lccc}
\toprule
\multirow{2}{*}{\textbf{Model}} & \multirow{2}{*}{\textbf{Top-1}} & \textbf{Params} &  \textbf{Search time}  \\
& &\textbf{(M)}   &\textbf{(GPU days)} \\ \hline
MobileNetV2 \cite{sandler2018mobilenetv2} & 72.0 & 3.4 & -  \\
ShuffleNetV2 2x (V2) \cite{ma2018shufflenet} & 73.7 & $\sim$5 & - \\ 
\hline
NASNet-A \cite{zoph2018learning} & 74.0 & 5.3 & 1800 \\
AmoebaNet-A \cite{real2018regularized} & 74.5 & 5.1 & 3150 \\
MnasNet-92 \cite{Tan_2019_CVPR} & 74.8 & 4.4 & -  \\
SNAS \cite{xie2018snas} & 72.7  & 4.3 & 522  \\
DARTS \cite{liu2018darts} & 73.1  & 4.9 & 4 \\ \hline
RL + BPE-1 (Ours)  & 74.18 & 5.5 & 0.33  \\
EA + BPE-1 (Ours) & 74.56 & 5.0 & 0.33  \\
RS + BPE-1 (Ours) & 74.2 & 5.5 & 0.33  \\
DARTS \cite{liu2018darts} + BPE-1 (Ours) & 74.0 & 5.9 & 0.05  \\
 \bottomrule
\end{tabular}}
\end{center}
\caption{Comparison with the state-of-the-art image classification methods on ImageNet. All the NAS networks in this table are searched on CIFAR10, and are then directly transferred to ImageNet.}
\label{tab:imagenet_results}
\end{table}

\subsubsection{Result on CIFAR10}
We compare our method with both manually designed networks and NAS networks. 
The manually designed networks include ResNet \cite{he2016deep}, DenseNet \cite{huang2017densely} and SENet \cite{hu2018squeeze}. 
We evaluate on four categories of NAS methods, \emph{i.e.}, RL methods (NASNet \cite{zoph2018learning}, ENAS \cite{pham2018efficient} and Path-level NAS \cite{cai2018path}), evolutional algorithms (AmoebaNet \cite{real2018regularized}), gradient-based methods (DARTS \cite{liu2018darts}) and Random Search.

The results for convolutional architectures on CIFAR-10 are presented in Tab.~\ref{tab:cifar_results}. 
It is worth noting that the found BPE combining with various search algorithms outperform various state-of-the-art search algorithms \cite{zoph2018learning,liu2018darts,real2019aging} in accuracy, with much lower computational consumption (only $0.05$ GPU days $\ll$ $3150$ in \cite{real2018regularized}). 
We attribute our superior results to the found BPE. 
Another notable observation from Tab.~\ref{tab:cifar_results} is that, even with random search in the search space, the test error rate is only $2.44$\%, which outperforms previous methods in the same search space. 
Conclusively, with the found BPE, search algorithms can quickly explore the architecture search space and generates a better architecture. 
We also report the results of hand-crafted networks in Tab.~\ref{tab:cifar_results}. 
Clearly, our method shows a notable enhancement.

\begin{figure}[t]
\includegraphics[width=1.0\linewidth]{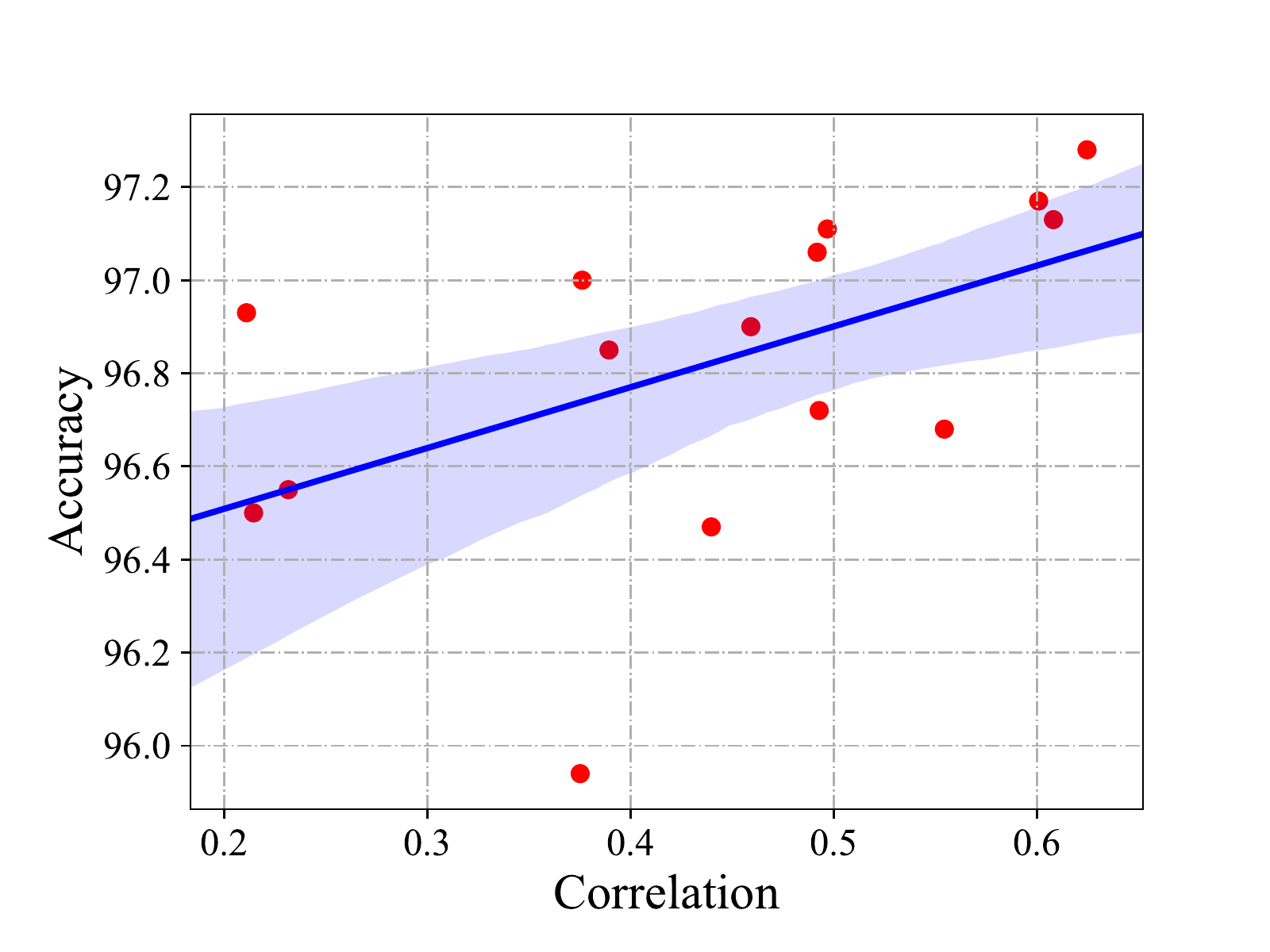}
\caption{The relationship between $r_s$ and performance with random sampled BPEs in $\mathcal{D}$. The x-axis measures the Spearman Rank Correlation $r_s$ by Eq.~\ref{eq:objective}, and the y-axis measures the real architecture performance found by DARTS+BPE on CIFAR10. The correlation between $r_s$ and performance is $0.65$.\label{fig:objective_effectiveness}}
\end{figure}
\begin{figure*}
\centering
\includegraphics[width=0.9\linewidth]{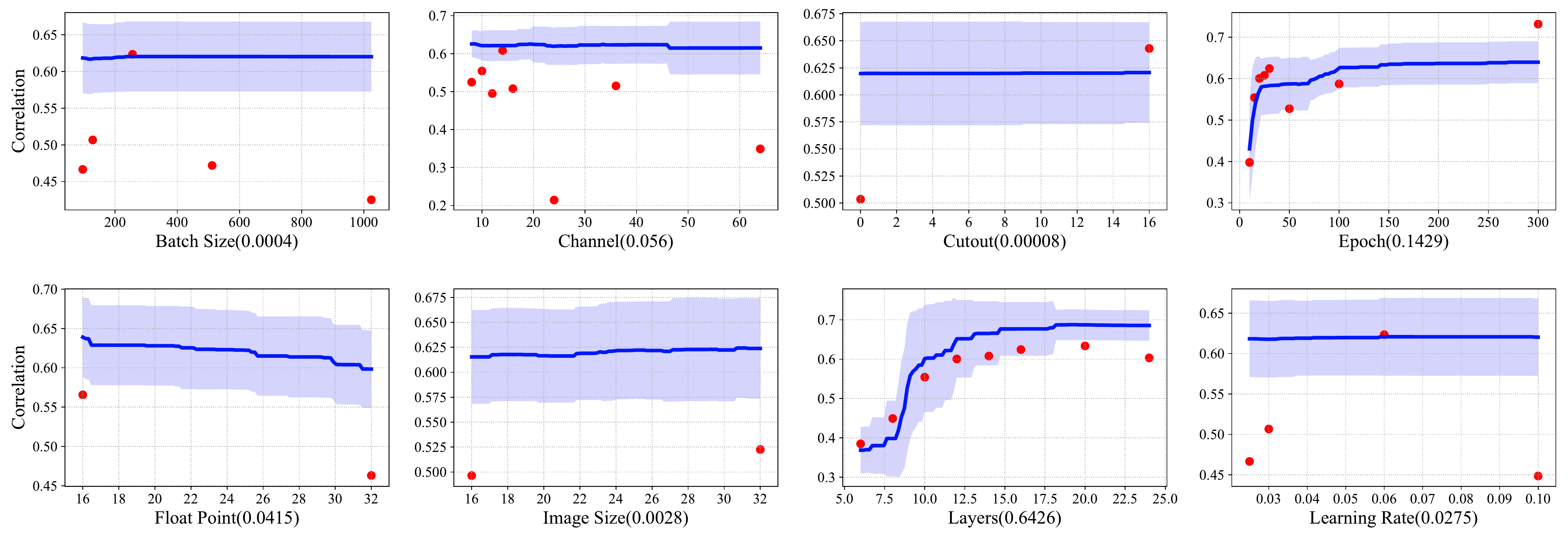}
\caption{The importance (within brackets) and regression predict curves with mean and variance learned by the random forest for each hyper-parameter. Importance is highly correlated to the curve steepnes. For the two most important parameters (epoch and layer), we can get a high $r_s$ within a small range.\label{fig:feature_importance}}
\end{figure*}

\subsubsection{Results on ImageNet}
We further compare our method under the mobile settings on ImageNet to demonstrate the generalizability. 
The best architecture on CIFAR-10 is transferred to ImageNet, which follows the same experimental settings in \cite{zoph2018learning,pham2018efficient,cai2018path}. 
Results in Tab.~\ref{tab:imagenet_results} show that the best cell architecture on CIFAR10 is transferable to ImageNet. 
The proposed method achieves comparable accuracy to the state-of-the-art methods \cite{zoph2018learning,real2018regularized,liu2018progressive,real2018regularized,liu2018progressive,pham2018efficient,liu2018darts,cai2018path} while using far less computational resources, \emph{e.g.}, $9,545$ times faster comparing to EA, and $5,400$ times faster comparing to RL.

\subsection{Deep Analysis in Performance Estimation} \label{sec:analysis}
We further study the efficiency of using Spearman Rank Correlation $r_s$ as the objective function in Fig.~\ref{fig:objective_effectiveness}. 
In Fig.~\ref{fig:feature_importance}, we also provide a deep analysis about the importance of every hyper-parameter. 
One can make full use of this analysis to transfer the found BPEs to other datasets and tasks.

We randomly select $15$ hyper-parameter settings in $\Omega$ and apply them on the DARTS \cite{liu2018darts} search algorithm to find optimal architectures. 
Fig.~\ref{fig:objective_effectiveness} illustrates the relationship between $r_s$ and the accuracy of the optimal architecture found by the corresponding setting. 
The performance is highly correlated to $r_s$ (with a $0.65$ correlation), which denotes the efficiency of the proposed objective function in Eq.~\ref{eq:objective}.

After exploring the BPE space by the proposed method, we get a dataset $\mathcal{D}$ \emph{w.r.t.} each $\Theta_i$ and $r_s$, which is used as the training set to train a random forest regression predictor \cite{hoos2014efficient} for each $\Theta_i$.
We then report the $r_s$ estimated by the predictor and importance for each hyper-parameter in Fig.~\ref{fig:feature_importance}. 
As illustrated in Fig.~\ref{fig:feature_importance}, the steepness and importance are highly correlated, \emph{i.e.,} the more important the parameter is, the steeper of the corresponding curve is, vice versa. 
At the same time, for the two most important parameters (epoch and layer), we get a high $r_s$ with a small range. 
This means that we only need to carefully finetune these two parameters in a small range when transferring to other datasets.

\begin{table}[]
\small
\begin{center}
\begin{tabular}{llcccc}
\toprule
\multicolumn{2}{l}{\multirow{2}{*}{}} & \multicolumn{4}{c}{Epoch}  \\ \cline{3-6} 
\multicolumn{2}{l}{}                  & 50 & 200 & 400 & 600 \\ \hline
\multirow{2}{*}{Fair}        & Global $r_s$& 0.10    & -0.06   & 0.13   & -0.03   \\ \cline{2-6} 
                             & Local $r_s$ & 0.13   & 0.50   & 0.31   & 0.31   \\ \hline
\multirow{2}{*}{Random}      & Global $r_s$& 0.10   & -0.05   & -0.30   & -0.19   \\ \cline{2-6} 
                             & Local $r_s$ & -0.14  & -0.52   & 0.61   & -0.01   \\ \hline
\multirow{2}{*}{Random\_10}  & Global $r_s$& 0.0   & 0.0   & 0.02   & -0.08   \\ \cline{2-6} 
                             & Local $r_s$ & 0.26  & 0.11   & 0.57   & 0.58   \\ 
                       \bottomrule
\end{tabular}
\end{center}
\caption{Comparison of the Global $r_s$ and Local $r_s$ under different training conditions. ``Fair" denotes each operation in an edge is trained with exactly the same epoch. ``Random" denotes each operation in an edge is trained randomly with different random level. Global $r_s$ and local $r_s$ denote we use the trained model to evaluate the performance estimation globally and locally, respectively.}
\label{tab:understanding}
\end{table}

\subsection{Understanding One-shot based Methods}\label{sec:unstanding_one_shot}
Previous works \cite{li2019random,sciuto2019evaluating} have reported that one-shot based methods such as DARTS do not work well (in some cases even no better than random search). 
There are two main questions which are not been explained yet: 
(1) One-shot based methods can not make a good estimate of performance, but they can search for good neural architectures. 
(2) The instability of one-shot based methods, that is, the found networks are different with different random seeds. 
With the found BPE, we can effectively investigate every search phase in these methods. 

To understand and explain such questions, we first train the same hypergraph with different settings: 
(1) Fair training, each operation in an edge is trained with exactly the same epoch; 
(2) Random training, each operation in an edge is trained randomly at different random levels. 
In Tab.~\ref{tab:understanding}, we report the global and local $r_s$ in the case of fair training and random training.  
The global $r_s$ denotes that we use our trained hypergraph to get the validation performance for the networks in $\mathcal{G}$, and then calculate the $r_s$ with $X_f$. 
The local $r_s$ is obtained by the following steps: When training the hypergraph, we save the sampled network architectures $\mathcal{G}_t$ and the corresponding validation performance $X_{\mathcal{G}_t}$ at epoch $t$. 
The local $r_s$ is then obtained by using the found BPE-2 and Eq.~\ref{eq:objective} \emph{i.e.,} $r_s(X_{\mathcal{G}_t}, $BPE-2$(\mathcal{G}_t))$. 
As illustrated in Tab.~\ref{tab:understanding}, one-shot based methods have a poor performance estimation in global $r_s$, which is consistent with previous works \cite{li2019random,sciuto2019evaluating}. 
However, these methods have a high local $r_s$, which means that these methods are essentially using the local information. 
That is to say, each epoch in the search phase can only perceive and optimize by using local information, which reasonably explains the instability of DARTS.

\section{Conlusion} \label{sec:6}
In this paper, we present the first systematic analysis of the budgeted performance estimation (BPE) in NAS, and propose a minimum importance pruning (MIP) towards optimal PE. The proposed MIP gradually reduces the number of BPE hyper-parameters, which allocates more computation resources on more important hyper-parameters. 
The found MIP-BPE is generalized to various search algorithms, including reinforcement learning, random search, evolution algorithm and gradient-based methods.
Combining the found BPE with various NAS algorithms, we have reached the state-of-the-art test error $2.66\%$ on CIFAR10 with much fewer search time, which also helps us to better understand the widely-used one-shot based methods.

\paragraph{Acknowledgements.}
\footnotesize{This work is supported by the Nature Science Foundation of China (No.U1705262, No.61772443, No.61572410, No.61802324 and No.61702136), 
National Key R\&D Program (No.2017YFC0113000, and No.2016YFB1001503), 
and Nature Science Foundation of Fujian Province, China (No. 2017J01125 and No. 2018J01106).}

{\small
\bibliographystyle{ieee_fullname}
\bibliography{egbib}

\begin{thebibliography}{10}\itemsep=-1pt

\bibitem{AkimotoICML2019}
Youhei Akimoto, Shinichi Shirakawa, Nozomu Yoshinari, Kento Uchida, Shota
  Saito, and Kouhei Nishida.
\newblock Adaptive stochastic natural gradient method for one-shot neural
  architecture search.
\newblock In {\em ICML}, 2019.

\bibitem{back1996evolutionary}
Thomas Back.
\newblock {\em Evolutionary algorithms in theory and practice: evolution
  strategies, evolutionary programming, genetic algorithms}.
\newblock Oxford university press, 1996.

\bibitem{bender2018understanding}
Gabriel Bender, Pieter-Jan Kindermans, Barret Zoph, Vijay Vasudevan, and Quoc
  Le.
\newblock Understanding and simplifying one-shot architecture search.
\newblock In {\em ICML}, 2018.

\bibitem{bergstra2012random}
James Bergstra and Yoshua Bengio.
\newblock Random search for hyper-parameter optimization.
\newblock {\em JMLR}, 2012.

\bibitem{bojarski2016end}
Mariusz Bojarski, Davide Del~Testa, Daniel Dworakowski, Bernhard Firner, Beat
  Flepp, Prasoon Goyal, Lawrence~D Jackel, Mathew Monfort, Urs Muller, Jiakai
  Zhang, et~al.
\newblock End to end learning for self-driving cars.
\newblock {\em arXiv}, 2016.

\bibitem{breiman2001random}
Leo Breiman.
\newblock Random forests.
\newblock {\em Machine learning}, 2001.

\bibitem{cai2018efficient}
Han Cai, Tianyao Chen, Weinan Zhang, Yong Yu, and Jun Wang.
\newblock Efficient architecture search by network transformation.
\newblock In {\em AAAI}, 2018.

\bibitem{cai2018path}
Han Cai, Jiacheng Yang, Weinan Zhang, Song Han, and Yong Yu.
\newblock Path-level network transformation for efficient architecture search.
\newblock {\em arXiv}, 2018.

\bibitem{chen2019binarized}
Hanlin Chen, Li'an Zhuo, Baochang Zhang, Xiawu Zheng, Jianzhuang Liu, David
  Doermann, and Rongrong Ji.
\newblock Binarized neural architecture search.
\newblock {\em arXiv preprint arXiv:1911.10862}, 2019.

\bibitem{chen2018searching}
Liang-Chieh Chen, Maxwell Collins, Yukun Zhu, George Papandreou, Barret Zoph,
  Florian Schroff, Hartwig Adam, and Jon Shlens.
\newblock Searching for efficient multi-scale architectures for dense image
  prediction.
\newblock In {\em NeurIPS}, 2018.

\bibitem{chen2019progressive}
Xin Chen, Lingxi Xie, Jun Wu, and Qi Tian.
\newblock Progressive differentiable architecture search: Bridging the depth
  gap between search and evaluation.
\newblock {\em ICCV}, 2019.

\bibitem{Cheng_2019_CVPR}
Hao Cheng, Dongze Lian, Bowen Deng, Shenghua Gao, Tao Tan, and Yanlin Geng.
\newblock Local to global learning: Gradually adding classes for training deep
  neural networks.
\newblock In {\em The IEEE Conference on Computer Vision and Pattern
  Recognition (CVPR)}, June 2019.

\bibitem{devries2017improved}
Terrance DeVries and Graham~W Taylor.
\newblock Improved regularization of convolutional neural networks with cutout.
\newblock {\em arXiv}, 2017.

\bibitem{Dong_2019_CVPR}
Xuanyi Dong and Yi Yang.
\newblock Searching for a robust neural architecture in four gpu hours.
\newblock In {\em CVPR}, 2019.

\bibitem{he2016deep}
Kaiming He, Xiangyu Zhang, Shaoqing Ren, and Jian Sun.
\newblock Deep residual learning for image recognition.
\newblock In {\em CVPR}, 2016.

\bibitem{hoos2014efficient}
Holger Hoos and Kevin Leyton-Brown.
\newblock An efficient approach for assessing hyperparameter importance.
\newblock In {\em ICML}, pages 754--762, 2014.

\bibitem{hu2018squeeze}
Jie Hu, Li Shen, and Gang Sun.
\newblock Squeeze-and-excitation networks.
\newblock In {\em CVPR}, 2018.

\bibitem{huang2017densely}
Gao Huang, Zhuang Liu, Laurens Van Der~Maaten, and Kilian~Q Weinberger.
\newblock Densely connected convolutional networks.
\newblock In {\em CVPR}, 2017.

\bibitem{hutter2011sequential}
Frank Hutter, Holger~H Hoos, and Kevin Leyton-Brown.
\newblock Sequential model-based optimization for general algorithm
  configuration.
\newblock In {\em LION}, 2011.

\bibitem{ji2019semi}
Rongrong Ji, Ke Li, Yan Wang, Xiaoshuai Sun, Feng Guo, Xiaowei Guo, Yongjian
  Wu, Feiyue Huang, and Jiebo Luo.
\newblock Semi-supervised adversarial monocular depth estimation.
\newblock {\em IEEE transactions on pattern analysis and machine intelligence},
  2019.

\bibitem{kaelbling1996reinforcement}
Leslie~Pack Kaelbling, Michael~L Littman, and Andrew~W Moore.
\newblock Reinforcement learning: A survey.
\newblock {\em JAIR}, 1996.

\bibitem{klein2016fast}
Aaron Klein, Stefan Falkner, Simon Bartels, Philipp Hennig, and Frank Hutter.
\newblock Fast bayesian optimization of machine learning hyperparameters on
  large datasets.
\newblock {\em arXiv}, 2016.

\bibitem{krizhevsky2010convolutional}
Alex Krizhevsky and Geoffrey Hinton.
\newblock Learning multiple layers of features from tiny images.
\newblock Technical report, 2009.

\bibitem{krizhevsky2009learning}
Alex Krizhevsky, Geoffrey Hinton, et~al.
\newblock Learning multiple layers of features from tiny images.
\newblock Technical report, 2009.

\bibitem{li2019random}
Liam Li and Ameet Talwalkar.
\newblock Random search and reproducibility for neural architecture search.
\newblock {\em arXiv}, 2019.

\bibitem{liu2019auto}
Chenxi Liu, Liang-Chieh Chen, Florian Schroff, Hartwig Adam, Wei Hua, Alan
  Yuille, and Li Fei-Fei.
\newblock Auto-deeplab: Hierarchical neural architecture search for semantic
  image segmentation.
\newblock {\em CVPR}, 2019.

\bibitem{liu2018progressive}
Chenxi Liu, Barret Zoph, Maxim Neumann, Jonathon Shlens, Wei Hua, Li-Jia Li, Li
  Fei-Fei, Alan Yuille, Jonathan Huang, and Kevin Murphy.
\newblock Progressive neural architecture search.
\newblock In {\em ECCV}, 2018.

\bibitem{liu2018darts}
Hanxiao Liu, Karen Simonyan, and Yiming Yang.
\newblock Darts: Differentiable architecture search.
\newblock {\em ICLR}, 2019.

\bibitem{ma2018shufflenet}
Ningning Ma, Xiangyu Zhang, Hai-Tao Zheng, and Jian Sun.
\newblock Shufflenet v2: Practical guidelines for efficient cnn architecture
  design.
\newblock In {\em ECCV}, 2018.

\bibitem{paszke2017automatic}
Adam Paszke, Sam Gross, Soumith Chintala, Gregory Chanan, Edward Yang, Zachary
  DeVito, Zeming Lin, Alban Desmaison, Luca Antiga, and Adam Lerer.
\newblock Automatic differentiation in pytorch.
\newblock 2017.

\bibitem{pham2018efficient}
Hieu Pham, Melody~Y Guan, Barret Zoph, Quoc~V Le, and Jeff Dean.
\newblock Efficient neural architecture search via parameter sharing.
\newblock {\em ICML}, 2018.

\bibitem{radosavovic2019network}
Ilija Radosavovic, Justin Johnson, Saining Xie, Wan-Yen Lo, and Piotr
  Doll{\'a}r.
\newblock On network design spaces for visual recognition.
\newblock {\em arXiv}, 2019.

\bibitem{real2019aging}
E Real, A Aggarwal, Y Huang, and QV Le.
\newblock Aging evolution for image classifier architecture search.
\newblock In {\em AAAI}, 2019.

\bibitem{real2018regularized}
Esteban Real, Alok Aggarwal, Yanping Huang, and Quoc~V Le.
\newblock Regularized evolution for image classifier architecture search.
\newblock {\em AAAI}, 2019.

\bibitem{russakovsky2015imagenet}
Olga Russakovsky, Jia Deng, Hao Su, Jonathan Krause, Sanjeev Satheesh, Sean Ma,
  Zhiheng Huang, Andrej Karpathy, Aditya Khosla, Michael Bernstein, et~al.
\newblock Imagenet large scale visual recognition challenge.
\newblock {\em IJCV}, 2015.

\bibitem{sandler2018mobilenetv2}
Mark Sandler, Andrew Howard, Menglong Zhu, Andrey Zhmoginov, and Liang-Chieh
  Chen.
\newblock Mobilenetv2: Inverted residuals and linear bottlenecks.
\newblock In {\em CVPR}, 2018.

\bibitem{saxena2016convolutional}
Shreyas Saxena and Jakob Verbeek.
\newblock Convolutional neural fabrics.
\newblock In {\em NeurlPS}, 2016.

\bibitem{sciuto2019evaluating}
Christian Sciuto, Kaicheng Yu, Martin Jaggi, Claudiu Musat, and Mathieu
  Salzmann.
\newblock Evaluating the search phase of neural architecture search.
\newblock {\em arXiv}, 2019.

\bibitem{snoek2012practical}
Jasper Snoek, Hugo Larochelle, and Ryan~P Adams.
\newblock Practical bayesian optimization of machine learning algorithms.
\newblock In {\em NeurIPS}, pages 2951--2959, 2012.

\bibitem{Tan_2019_CVPR}
Mingxing Tan, Bo Chen, Ruoming Pang, Vijay Vasudevan, Mark Sandler, Andrew
  Howard, and Quoc~V. Le.
\newblock Mnasnet: Platform-aware neural architecture search for mobile.
\newblock In {\em CVPR}, 2019.

\bibitem{thornton2013auto}
Chris Thornton, Frank Hutter, Holger~H Hoos, and Kevin Leyton-Brown.
\newblock Auto-weka: Combined selection and hyperparameter optimization of
  classification algorithms.
\newblock In {\em SIGKDD}, 2013.

\bibitem{CMIL2019}
Fang Wan, Chang Liu, Wei Ke, Xiangyang Ji, Jianbin Jiao, and Qixiang Ye.
\newblock C-mil: Continuation multiple instance learning for weakly supervised
  object detection.
\newblock In {\em IEEE CVPR}, pages 2199--2208, 2019.

\bibitem{xie2018snas}
Sirui Xie, Hehui Zheng, Chunxiao Liu, and Liang Lin.
\newblock {SNAS}: stochastic neural architecture search.
\newblock In {\em ICLR}, 2019.

\bibitem{xu2019pc}
Yuhui Xu, Lingxi Xie, Xiaopeng Zhang, Xin Chen, Guo-Jun Qi, Qi Tian, and
  Hongkai Xiong.
\newblock Pc-darts: Partial channel connections for memory-efficient
  differentiable architecture search.
\newblock {\em arXiv}, 2019.

\bibitem{zela2018towards}
Arber Zela, Aaron Klein, Stefan Falkner, and Frank Hutter.
\newblock Towards automated deep learning: Efficient joint neural architecture
  and hyperparameter search.
\newblock {\em ICML Workshop}, 2018.

\bibitem{FreeAnchor2019}
Xiaosong Zhang, Fang Wan, Chang Liu, Rongrong Ji, and Qixiang Ye.
\newblock Freeanchor: Learning to match anchors for visual object detection.
\newblock In {\em NeurIPS}, pages 147--155, 2019.

\bibitem{zheng2018centralized}
Xiawu Zheng, Rongrong Ji, Xiaoshuai Sun, Yongjian Wu, Feiyue Huang, and Yanhua
  Yang.
\newblock Centralized ranking loss with weakly supervised localization for
  fine-grained object retrieval.
\newblock In {\em IJCAI}, pages 1226--1233, 2018.

\bibitem{zheng2019towards}
Xiawu Zheng, Rongrong Ji, Xiaoshuai Sun, Baochang Zhang, Yongjian Wu, and
  Feiyue Huang.
\newblock Towards optimal fine grained retrieval via decorrelated centralized
  loss with normalize-scale layer.
\newblock In {\em Proceedings of the AAAI Conference on Artificial
  Intelligence}, volume~33, pages 9291--9298, 2019.

\bibitem{zheng2019dynamic}
Xiawu Zheng, Rongrong Ji, Lang Tang, Yan Wan, Baochang Zhang, Yongjian Wu,
  Yunsheng Wu, and Ling Shao.
\newblock Dynamic distribution pruning for efficient network architecture
  search.
\newblock {\em arXiv}, 2019.

\bibitem{Zheng_2019_ICCV}
Xiawu Zheng, Rongrong Ji, Lang Tang, Baochang Zhang, Jianzhuang Liu, and Qi
  Tian.
\newblock Multinomial distribution learning for effective neural architecture
  search.
\newblock In {\em ICCV}, 2019.

\bibitem{zoph2016neural}
Barret Zoph and Quoc~V Le.
\newblock Neural architecture search with reinforcement learning.
\newblock {\em arXiv}, 2016.

\bibitem{zoph2018learning}
Barret Zoph, Vijay Vasudevan, Jonathon Shlens, and Quoc~V Le.
\newblock Learning transferable architectures for scalable image recognition.
\newblock In {\em CVPR}, 2018.

\end{thebibliography}
}
\end{document}